\documentclass[11pt]{article}
\RequirePackage{rotating} 

\usepackage{boxedminipage}
\usepackage[vlined, ruled]{algorithm2e}

\usepackage{titlesec}
\usepackage[utf8]{inputenc}
\usepackage{caption, makecell}

\usepackage{threeparttable}

\usepackage{geometry}
\usepackage{mdframed}
\usepackage{booktabs}

\titleformat{\paragraph}
{\normalfont\normalsize\itshape}{\theparagraph}{1em}{}
\titlespacing*{\paragraph}
{0pt}{3.25ex plus 1ex minus .2ex}{1.5ex plus .2ex}

\usepackage{graphicx}
\usepackage{mathrsfs,lscape,subfigure,enumerate,epstopdf}
\usepackage{amssymb}
\usepackage{pifont}
\usepackage{amsmath,lipsum}
\usepackage{amsthm}
\usepackage{enumerate} 
\usepackage{pgfplots}
\usepackage{tikz}
\usepackage{bm}
\usepackage{float}

\pgfplotsset{width=10cm,compat=1.16}
\usetikzlibrary{shapes.geometric, arrows, calc, fit, positioning, chains, arrows.meta,decorations.pathreplacing,matrix,backgrounds,positioning,shapes}

\usepackage{lineno}
\usepackage{longtable}
\usepackage{threeparttable}
\usepackage{multicol}
\usepackage{multirow}
\usepackage{booktabs}
\usepackage{diagbox}
\usepackage{natbib}
\setcitestyle{square,numbers}
\usepackage{fancybox}
\usepackage{textcomp}
\usepackage{rotating}
\usepackage{pdflscape} 

\usepackage{colortbl}
\usepackage{tabulary}
\usepackage{etoolbox}
\usetikzlibrary{patterns}
\usepackage{afterpage,lscape}
\usetikzlibrary{decorations.pathreplacing,calligraphy}
\usepackage[colorlinks]{hyperref}
\usepackage{authblk}

\usepackage{mathrsfs}

\usepackage[titletoc,title]{appendix}

\title{Dual Policy Reinforcement Learning for Real-time Rebalancing in Bike-sharing Systems}

\author[1,4,5,6]{Jiaqi Liang}

\author[1,4]{Defeng Liu}

\author[2,6]{Sanjay Dominik Jena}

\author[3,4]{Andrea Lodi}

\author[1,5,6]{Thibaut Vidal}

\affil[1]{\small Department of Mathematics and Industrial Engineering, Polytechnique Montréal
}

\affil[2]{\small School of Management, Université du Québec à Montréal
}

\affil[3]{\small Cornell Tech and Technion -- IIT 
}
\affil[4]{\small Canada Excellence Research Chair in Data-Science for Real-time Decision-Making (CERC)}

\affil[5]{\small SCALE-AI Chair in Data-Driven Supply Chains
}

\affil[6]{\small Centre Interuniversitaire de Recherche sur les Reseaux d’Entreprise, la Logistique et le Transport (CIRRELT)}

\date{\vspace{-5ex}}

\begin{document}
\maketitle

\begin{abstract}
Bike-sharing systems play a crucial role in easing traffic congestion and promoting healthier lifestyles. However, ensuring their reliability and user acceptance requires effective strategies for rebalancing bikes. This study introduces a novel approach to address the real-time rebalancing problem with a fleet of vehicles. It employs a dual policy reinforcement learning algorithm that decouples inventory and routing decisions, enhancing realism and efficiency compared to previous methods where both decisions were made simultaneously. We first formulate the inventory and routing subproblems as a multi-agent Markov Decision Process within a continuous time framework. Subsequently, we propose a DQN-based dual policy framework to jointly estimate the value functions, minimizing the lost demand. To facilitate learning, a comprehensive simulator is applied to operate under a first-arrive-first-serve rule, which enables the computation of immediate rewards across diverse demand scenarios. We conduct extensive experiments on various datasets generated from historical real-world data, affected by both temporal and weather factors. Our proposed algorithm demonstrates significant performance improvements over previous baseline methods. It offers valuable practical insights for operators and further explores the incorporation of reinforcement learning into real-world dynamic programming problems, paving the way for more intelligent and robust urban mobility solutions.
\end{abstract}

\section{Introduction}
\label{sec:intro}

The development of Bike-Sharing Systems (BSS) has been instrumental in alleviating urban traffic congestion and reducing CO$_2$ emissions \citep{meddin2022}. However, the stochastic nature of user behavior and peak hour demands often lead to unbalanced stations, necessitating efficient rebalancing strategies to minimize lost demand.

\begin{figure}[!htbp]
\centering
\includegraphics[scale=0.48]{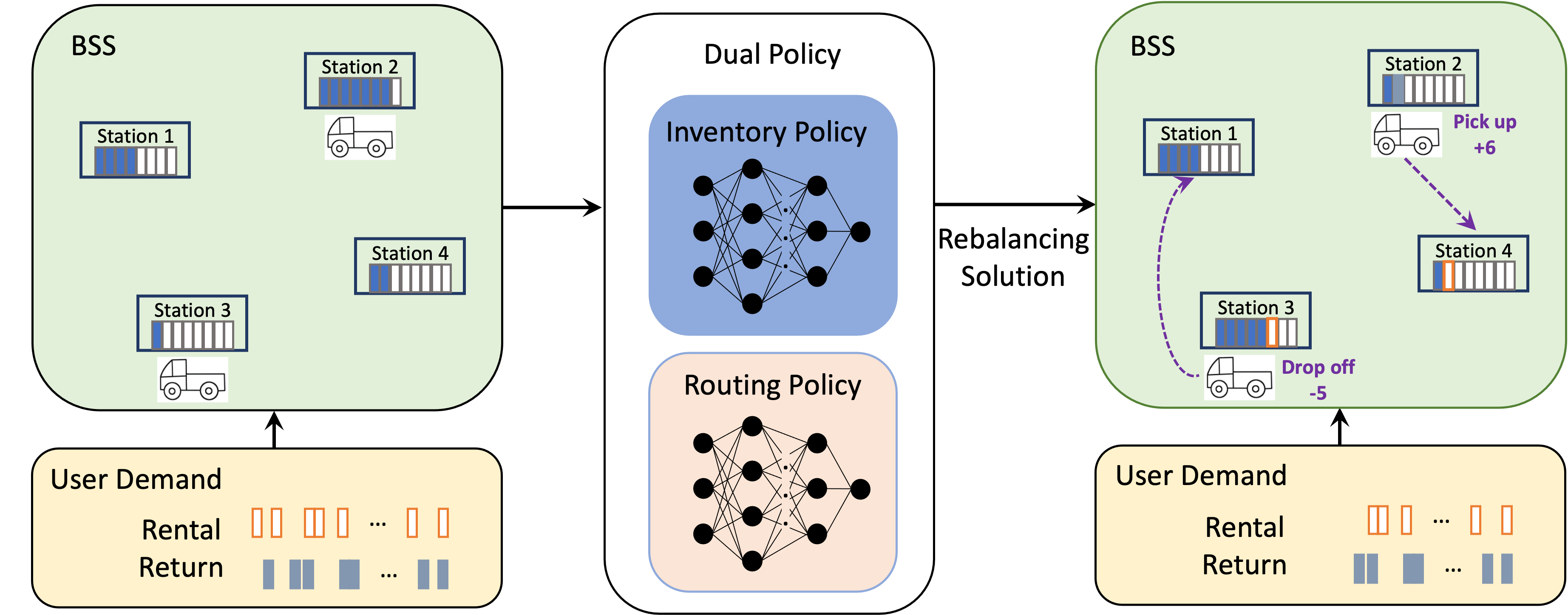}
\caption{Dynamic rebalancing in BSS and our dual policy approach: Station location and inventory information, vehicle location and inventory level, and user demand typically serve as inputs in dynamic rebalancing models. Here, we employ a dual policy to obtain rebalancing solution based on the environmental interactions among stations, vehicles, and users.}
\label{fig:Bss}
\end{figure}

The Dynamic Bike Repositioning Problem (DBRP) addresses the real-time redistribution of bikes across a network of stations to meet fluctuating demand during the day \citep{raviv2013static}. Typically, operators employ a fleet of vehicles to redistribute bikes among stations, aiming to minimize unmet demand. This task involves making both inventory decisions (how many bikes to pick up or drop off) and routing decisions (which station to visit next). The left part of Figure~\ref{fig:Bss} illustrates a simple BSS example with four stations and a fleet of two vehicles.

Markov Decision Processes (MDPs) have recently been harnessed to solve the DBRP in a few previous studies \citep[see, e.g.,][]{brinkmann2019dynamic, luo2022dynamic, seo2020dynamic} since they can capture the sequential nature of BSS decisions. In contrast, Mixed Integer Programming (MIP) models based on time discretizations are prevalent \citep[]{ghosh2019improving, liang2024dynamic}, but must strike a difficult balance between discretization accuracy and computational efficiency. Reinforcement Learning (RL) techniques offer promise in solving MDP models for DBRP. They learn optimal policies directly from interactions with the environment without requiring an explicit model of its dynamics, but their current applicability is often limited to single-vehicle rebalancing and small station networks due to high-dimensional action spaces. Moreover, existing MDP models simultaneously make inventory and routing decisions upon a vehicle's arrival at a station. By the time the inventory decision is completed, the system state has changed, potentially making the routing decision upon the vehicle's arrival suboptimal. Since system dynamics evolve during inventory rebalancing operations, decoupling inventory and routing decisions is critical.

This study proposes a spatio-temporal RL algorithm for DBRP with multiple vehicles under a continuous time framework, addressing the limitation of current MIP vehicle coordination methods in BSS that discretize time. We introduce a dual policy to decouple inventory and routing subproblems. This permits us to capture the dynamics of the environment more realistically and minimize lost demand. As shown in Figure~\ref{fig:Bss}, two Deep Q-networks (DQNs) are trained jointly for the inventory and routing subproblems to approximate the Q-value function. This approach ensures that system changes during inventory rebalancing are considered granularly rather than making inventory and routing decisions simultaneously as in previous RL models. To facilitate the learning process, we use a real-time BSS simulator to evaluate immediate rewards based on the BSS environment and various user demand scenarios. We conduct extensive experiments, including comparisons with MIP models and RL models with single policy or simultaneous inventory and routing decision-making, as well as ablation studies. The Dual Policy RL algorithm (DPRL) we propose significantly outperforms all benchmarks, demonstrating the effectiveness of DPRL for real-time rebalancing applications. Since other related problems, such as pickup-and-delivery and ride-sharing, share many common traits with the DBRP, involving both inventory and routing decisions, our approach will also serve as a guide and example for other supply chain and transportation optimization tasks beyond BSS.

The remainder of the paper is organized as follows. Section \ref{sec:literature} reviews related literature. Section \ref{sec:background} models DBRP as a Multi-agent Markov Decision Process (MMDP), while Section \ref{sec:method} presents our dual policy RL method for real-time rebalancing solutions. Section \ref{sec:experiments} provides numerical experiments and analyses, as well as an ablation study. Finally, Section~\ref{sec:conclusion} summarizes this work and outlines future research directions.

\section{Related Work}
\label{sec:literature}

Most existing works model DBRP using MIP models \citep[][etc]{ghosh2019improving, mellou2019dynamic,liang2024dynamic}. To remain tractable, MIP models require to simplify the problem, discretizing the planning horizon and assuming a constant number of station visits for each vehicle during a time period. Unfortunately, such limitations are unrealistic and can hinder the BSS from fully optimizing vehicle coordination and rebalancing.

MDP, as a sequential decision-making tool, has been increasingly applied to DBRP in BSS and offers a meaningful alternative to MIP methods. Despite its potential, MDP adoption in DBRP still remains nascent and hindered by its computational complexity. \cite{brinkmann2019dynamic} and \cite{brinkmann2020multi} pioneered MDP-based approaches, introducing dynamic lookahead policies for single and multiple vehicles, respectively. \citep{legros2019dynamic} developed a decision-support tool aiming to minimize unsatisfied users, employing a one-step policy improvement method, while \cite{luo2022dynamic} proposed a policy approximation algorithm for single-vehicle rebalancing. The evolving application of MDP in DBRP underscores the need for research addressing computational complexity and vehicle synchronization for enhanced decision-making efficiency. 

RL trains agents to make sequential decisions by interacting with environments and learning optimal strategies through trial and error, potentially solving MDP models. RL's success in complex optimization tasks motivates its application in BSS, where the complexity is akin to that in gaming, finance, and combinatorial optimization \citep{sutton2018reinforcement, oroojlooyjadid2019review, bello2016neural, bengio2021machine, Liu_Fischetti_Lodi_2022}.

Dynamic rebalancing involves two key decisions: the station a vehicle should visit next (routing decision), and the number of bikes to pick up or drop off upon a vehicle's arrival (inventory decision). \cite{li2018dynamic} proposed a clustering algorithm to group stations and then introduced an RL model to learn the rebalancing policy inside each cluster. \cite{xiao2018distributed} proposed a distributed RL solution with transfer learning, focusing only on a single policy (inventory decision). This approach allocates a separate agent to each station, responsible for exploring the best rebalancing strategy for its respective station. \cite{yindeep} proposed an RL-based rebalancing model, using DQN with a simulator under time discretization, disregarding the sequence of rentals, returns, and rebalancing operations. \cite{liang2024reinforcement} presented an RL method for DBRP with multiple vehicles where the inventory and routing decisions are made at the same time.
\cite{luo2022dynamic} and \cite{seo2022rebalancing} focused on the DBRP with a single vehicle, while \cite{li2018dynamic} considered multiple vehicles rebalancing in smaller clusters with fewer than 30 stations. 

Existing literature either focuses solely on inventory decisions or makes inventory and routing decisions simultaneously. The former overlooks the importance of routing decisions in dynamic rebalancing, while the latter ignores system state changes during inventory rebalancing operations, leading to suboptimal routing decisions. Moreover, the framework needs to be carefully considered when making simultaneous inventory and routing decisions.  These decisions can be made either upon the vehicle's arrival (inventory decision for the arriving station) or upon the vehicle's departure (inventory and routing decisions both for the next station).

To address these limitations, we propose a novel dual-policy RL approach that decouples inventory and routing decisions. Specifically, upon arrival at a station, the inventory policy determines the number of bikes to pick up or drop off. Once the inventory operation is completed, the routing policy decides the next station to visit. This separation enhances the realism and responsiveness of the system to dynamic changes, offering a more nuanced and effective solution to the real-time rebalancing problem.

\section{Problem Formulation}
\label{sec:background}

\subsection{The Dynamic Bike Repositioning Problem (DBRP)}

We consider a BSS with $|N|$ stations and a fleet $V$ of vehicles tasked with rebalancing bikes among stations. Each station $n \in N$ has a capacity $C_{n}$ of docks, and each vehicle $v \in V$ has a capacity $\hat{C}_{v}$. The initial system state is characterized by the bike inventory $d^{n}_{0}$ at each station $n \in N$, along with the inventory ${p}^{v}_{0}$ and location ${z}^{v}_{0}$ of each vehicle $v \in V$. Distances and transit times between stations $i \in N$ and $j \in N$ are represented by $D_{i,j}$ and $R_{i,j}$, respectively. The estimated time to load or unload one bike is denoted by $\beta$ minutes. Lost demand occurs when rental or return requests cannot be fulfilled due to bike unavailability or lack of available docks, respectively. The primary goal is optimizing the rebalancing strategies for vehicles across the station network, minimizing lost demand. A complete list of problem parameters is summarized in Table ~\ref{defin of para3}.

\textbf{\begin{table}[!tbph]
\caption{Parameters of DBRP}
\label{defin of para3}
\centering
\scalebox{0.85}{
\begin{tabular}{ll}
\hline
Parameter              & Definition                                                                                        \\ \hline
$N$                    & Set of stations                                                                               \\ 
$V$                    & Set of vehicles                                                                               \\
$C_{n}$       & Capacity of station $n \in N$                                                                      \\
$\hat{C}_{v}$ & Capacity of vehicle $v \in V$                                                                       \\ 
$D_{i,j}$              & Distance between station $i\in N$ and $j\in N$                                                          \\
$R_{i,j}$              & Transit time between station $i\in N$ and $j\in N$                                                          \\
$\beta$              & Time (in minutes) for loading/unloading one bike                                                            \\ 
$d^{n}_{0}$   & Initial number of bikes in station $n \in N$                                      \\ 
${p}^{v}_{0}$ & Initial number of bikes in vehicle $v \in V$   \\ 
$z^{v}_{0}$   & Initial location (station) of each vehicle $v \in V$                                     \\ \hline
\end{tabular}}
\end{table}}

\subsection{Multi-agent Markov Decision Process (MMDP) Formulation}
\label{mmdp defin}

We formulate the DBRP as an MMDP, characterized by the tuple $(\bm{S}, \bm{A}, W, R, \gamma)$, where $\bm{S}$ is the set of system states, $\bm{A}$ is the set of actions for the agents, $W$ represents state transition probabilities, $R$ is the cumulative discounted total reward, and $\gamma$ is the discount factor that emphasizes short-term over long-term rewards. Agents operate through a sequence of steps, where each action leads to a new state and associated reward. The related notations are summarized in Table~\ref{defin of var3}.

\begin{table}[!tbph]
\caption{Notation of MMDP model for rebalancing}
\label{defin of var3}
\centering
\scalebox{0.85}{
\begin{tabular}{ll}
\hline
Symbol & Definition                                                                                                   \\ \hline
$K$                    & Sequence of steps                                                                                    \\ 
$\bm{S}$                    & Set of states                                                                                    \\ 
$T$                    & Time of each step, $T=\{t_{1},...,t_{|K|}\}$                                                                                               \\
$d_{k}^{n}$   & Number of bikes available at station $n \in N$ at step $k \in K$           \\
$b_{k}^{v}$ & Station that vehicle $v$ is or just departed from at step $k$                        \\ 
$g_{k}^{v}$ & Station that vehicle $v$ is or is heading to at step $k$                \\  
$i_{k}^{v}$   & Indicator, 1 indicates routing state; 0, inventory state  \\
$p_{k}^{v}$      & Inventory of vehicle $v$ at time $t_k$  \\ 
$m_{k}^{v}$      & Estimated time of vehicle $v$ to make the next decision \\ 
$o_{k}^{v}$      & Remaining inventory operations for $v$ at $t_k$ \\
$\bm{A}$                    & The set of actions                                                                              \\ 
$l_{k}^{v}$         & Inventory decision for vehicle $v$ at step $k$\\ 
$z_{k}^{v}$         & Routing decision for vehicle $v$ at step $k$\\ 
$\Pi$         & Set of policies\\
$\gamma $   & Discount factor\\
\hline
\end{tabular}}
\end{table}

\textbf{State Space.}
At each step $k \in K$, state $\bm{S_{k}} \in \bm{S}$ encapsulates the comprehensive system status, including time, station, and vehicle information. We denote $\bm{S_{k}}=(t_{k}, \bm{d_{k}}, \bm{H_{k}})$, where $t_{k}$ is the current time and $\bm{d_{k}}=(d_{k}^{n}, \forall n \in N)$ represents the inventory of each station. The vehicle status is denoted as $\bm{H_{k}}=(b_{k}^{v}, g_{k}^{v}, p_{k}^{v}, m_{k}^{v}, o_{k}^{v}, i_{k}^{v}, \forall v \in V)$, where $b_{k}^{v}$ signifies the current station at which vehicle $v$ is located or has just departed from, and $g_{k}^{v}$ indicates the current station or the next station that vehicle $v$ is traveling to. Note that $b_{k}^{v}$ and $g_{k}^{v}$ can be the same station when vehicle $v$ is performing inventory operations. $p_{k}^{v}$ denotes the current inventory of vehicle $v$, and $m_{k}^{v}$ is the estimated time for vehicle $v$ to make the next decision, either after completing the current inventory operation or upon arrival at the next station. The number of remaining rebalancing operations (i.e., drop-offs or pick-ups) is indicated by $o_{k}^{v}$. A binary indicator $i_{k}^{v}$ is introduced to facilitate the decoupling of inventory and routing decisions: $i_{k}^{v} = 0$ when vehicle $v$ arrives at a station and makes an inventory decision, and $i_{k}^{v} = 1$ if vehicle $v$ completes inventory operations and proceeds to make a routing decision. 

\textbf{Action Space.}
At each step $k \in K$, an action $\bm{a_{k}} \in \bm{A}$ is selected, representing either an inventory decision or a routing decision. For inventory decisions, let $l_{k}^{v}$ represent the action, and for routing decisions, let $z_{k}^{v}$ represent the action. To simplify inventory decision, we consider three predefined fill levels $\mu_{i} \in [0,1], \forall i \in \{1, 2, 3\}$, representing proportions of the station capacity.  Vehicles aim to adjust station inventory to one of these levels: $\smash{\mu_{1}C_{g^{v}_{k}}}$, $\smash{\mu_{2}C_{g^{v}_{k}}}$, or $\smash{\mu_{3}C_{g^{v}_{k}}}$. The feasible inventory decisions depend on the vehicle’s capacity $\smash{\hat{C}_{v}}$, the current vehicle inventory $p^{v}_{k}$, and the station inventory $d^{g^{v}_{k}}_{k}$. Thus, $l_{k}^{v}$ for vehicle $v$ at step $k$ are defined as follows:
\begin{equation}
(l_{k}^{v})_{i} =
\begin{cases}
 \min \{\hat{C}_{v}-p^{v}_{k}, d^{g^{v}_{k}}_{k}-\mu_{i}C_{g^{v}_{k}}\}& \text{if}\quad \mu_{i}C_{g^{v}_{k}}< d^{g^{v}_{k}}_{k} \\ 
 \max \{-p^{v}_{k}, d^{g^{v}_{k}}_{k}-\mu_{i}C_{g^{v}_{k}}\}& \text{if}\quad  \mu_{i}C_{g^{v}_{k}}> d^{g^{v}_{k}}_{k}\\ 
 0& \text{otherwise}.
\end{cases} \label{loading}
\end{equation} 

In the first case of Equation \eqref{loading}, $l_{k}^{v} > 0$, indicating the vehicle needs to pick up $l_{k}^{v}$ bikes from the station. In the second case, $l_{k}^{v} < 0$, indicating the vehicle needs to drop off $|l_{k}^{v}|$ bikes at the station. In the third case, no rebalancing operations are performed. The routing decision is given by $z_{k}^{v}$, indicating the next station that vehicle $v \in V$ will visit. A constraint is applied to ensure that no two vehicles are simultaneously directed towards the same station.

DBRP is here formulated such that an action is made only for a specific vehicle at a particular point when it arrives or departs from a station, allowing a smaller action space where vehicles can still take actions independently, taking account of other vehicles' status. This ensures that decision-making is in real-time and continuously adapts to global status changes in the system.

\textbf{Rewards and Returns.}
A fine-grained simulator 
is employed to capture the immediate reward $r_{k+1}$, which is the negative value of lost demand across all stations occurring between two states. The expected cumulative discounted return, $\smash{R_{k}= \mathbb{E} [\sum_{j=0}^{K-k-1}\gamma^{j}r_{k+j+1})]}$, reflects the long-term aggregated reward from a series of actions. The discount factor $\gamma \in [0,1]$ progressively reduces the value of future rewards as step $j$ increases. A planning solution is encoded through a policy $\pi \in \Pi$, where $\Pi$ denotes the set of all possible policies. A policy, represented as $\pi(\bm{a_{k}}|\bm{S_{k}})$, serves as a strategy dictating the probability of taking an action $a$ given state $s$. The overarching objective is to identify the optimal policy $\pi^{*}$ that maximizes the total expected reward over time. 

\section{Dual Policy Reinforcement Learning}
\label{sec:method}

This section introduces our DPRL framework, which decouples inventory and routing decisions for real-time rebalancing. This separation allows for more precise and responsive adjustments to system dynamics, significantly reducing lost demand and enhancing user satisfaction. We begin by presenting the tailored MMDP for the proposed dual policy framework, followed by the learning pipeline of DPRL.

\subsection{Tailored MMDP for Dual Policy}
\label{sec:talored_MMDP}

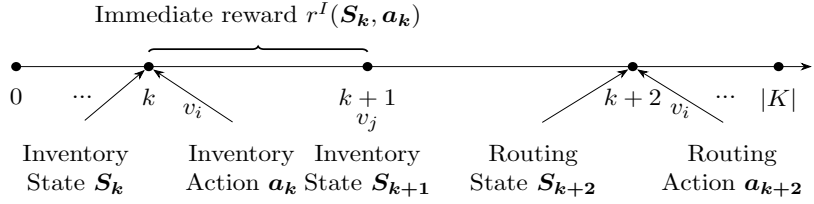
\begin{figure}[!hbtp]
\centering
  \noindent\makebox[\textwidth]{
  \resizebox{11cm}{2.8cm}{\begin{tikzpicture}[>=Stealth,
                    node distance=2cm,
                    on grid,
                    auto,
                    every node/.style={align=center}]

   \draw[->] (0,0) -- (12,0) node[anchor=north] {};

  \node[circle,fill,inner sep=1.5pt,label={[below, yshift=-0.3cm]{$0$}}] (start) at (0,0) {};
  \node[circle,fill,inner sep=1.5pt,label={[below, yshift=-0.3cm]{$k$}}] (k) at (2,0) {};
  \node[circle,fill,inner sep=1.5pt,label={[below, yshift=-0.3cm]{$k+1$}}] (k1) at (5.3,0) {};
  \node[circle,fill,inner sep=1.5pt,label={[below, yshift=-0.3cm]{$k+2$}}] (k2) at (9.3,0) {};
  \node[circle,fill,inner sep=1.5pt,label={[below, yshift=-0.3cm]{$|K|$}}] (end) at (11.5,0) {};

  \node[label=below:{$...$}] at (1,-0.2) {};
  \node[label=below:{Inventory \\State $\bm{S_k}$}] (sk) at (0.9,-1) {};
  \node[label=below:{Inventory \\Action $\bm{a_k}$}] (ak) at (3.4,-1) {}; 
  \node[label=below:{Inventory \\ State $\bm{S_{k+1}}$}] at (5.3,-1) {};
\node[label=below:{$v_j$}] at (5.3,-0.5) {};
 \node[label=below:{$...$}] at (10.7,-0.2) {};
 
 \node[label=below:{Routing \\State $\bm{S_{k+2}}$}] (sk2) at (7.8,-1) {};
  \node[label=below:{Routing \\Action $\bm{a_{k+2}}$}] (ak2) at (10.8,-1) {}; 

  \draw[->] (sk) -- (k) node[midway,below] {};
  \draw[->] (ak) -- (k) node[midway,below] {$v_i$};
  \draw[->] (sk2) -- (k2) node[midway,below] {};
  \draw[->] (ak2) -- (k2) node[midway,below] {$v_i$};

\draw [thick,decorate,decoration = {brace,raise=4pt}] (2,0.1) --  (5.3,0.1) node[midway, yshift= 1em]{Immediate reward $r^{I}(\bm{S_k}, \bm{a_k})$};
\end{tikzpicture}}
  }
\caption{Dual policy framework}
\label{fig: timef}
\end{figure}

Figure~\ref{fig: timef} illustrates the states and actions within our continuous-time dual policy framework. Two distinct state types exist: inventory state and routing state, sharing identical components as described in Section~\ref{mmdp defin}. A step corresponds to the moment a vehicle arrives at or departs from a station after completing rebalancing operations. For instance, an inventory state $\bm{S_{k}}$ occurs when vehicle $v_{i}$ arrives at station $g_{k}^{v_{i}}$, while all the other vehicles either rebalance bikes at their stations or relocate to their next stations. At this point, an action $l_{k}^{v_{i}}$ is generated, dictating the number of bikes for $v_{i}$ to load or unload at step $k$. The system then transitions to the next state $\bm{S_{k+1}}$, which may represent an inventory state for $v_{j}$ (as illustrated in Figure~\ref{fig: timef}) or a routing state for other vehicles. After vehicle $v_{i}$ completes the inventory rebalancing operation generated in inventory state $\bm{S_{k}}$, a routing decision is made before it departs from the station in routing state $\bm{S_{k+2}}$. 

In our dual policy framework, the immediate reward is computed as the negative value of lost rental and return demand occurring within the time segment between two consecutive states of the same type. Here, an example of inventory reward $r^{I}(\bm{S_k}, \bm{a_k})$ is shown for the duration $[t_k, t_{k+1}]$. If the state $\bm{S_{k-1}}$ is a routing state, then the corresponding routing reward is denoted as $r^{R}(\bm{S_{k-1}}, \bm{a_{k-1}})$ for the interval $[t_{k-1}, t_{k+2}]$.

In previous RL frameworks\citep[][etc]{li2018dynamic, yindeep, liang2024reinforcement}, inventory and routing decisions for vehicle $v_{i}$ are made simultaneously in state $\bm{S_{k}}$. However, when vehicle $v_{i}$ completes the inventory operation at $t_{k+2}$, the routing decision made at $t_{k}$ may no longer be optimal due to system changes, particularly user demand, that occur during the interval $[t_k, t_{k+2}]$. Our dual policy framework addresses this by decoupling inventory and routing decisions, considering system changes in a more realistic and granular manner. Therefore, our approach focuses on the inventory decision upon the vehicle's arrival at the station and defers routing decisions until after the current inventory operation is completed. This allows us to consider the latest system state before deciding on the next station to visit. Similarly, routing decisions are made upon departure from the station, and inventory decisions are deferred until arrival at the next station. This dual policy framework ensures that decisions are informed by real-time global system state, thereby enhancing the overall effectiveness and responsiveness of rebalancing strategies.

\subsection{DPRL Pipeline}
\label{rlframe}

\begin{figure}[!hbtp]
\centering
  \noindent\makebox[\textwidth]{
  \resizebox{10cm}{4.5cm}{\begin{tikzpicture}[every node/.style={font=\Large}, 
  block/.style={
    rectangle, 
    draw, 
    fill=white,
    align=center, 
    minimum width=2.5cm, 
    minimum height=1.2cm, 
    thick
  },
  decision/.style={
    diamond,
    draw,
    fill=white,
    align=center,
    aspect=2,
    minimum width=2.5cm,
    minimum height=1.2cm,
    thick
  },
  line/.style={
     -{Latex[scale=1.2]},  
    thick
  },
  background/.style={
    rectangle,
    rounded corners = 10pt,
    inner sep=0.4cm,
    draw,
    thick,
    fill=gray!20 
  }
]

\node[block] (DemandScenarios) at (0,0) {Demand\\ Scenarios};
\node[block, right=2.1cm of DemandScenarios] (Simulator) {Simulator};
\node[decision, right=0cm of Simulator,  yshift=-3cm] (Decision) {State \\ Type};
\node[block, right=4.2cm of Simulator, yshift=0.4cm] (InventoryPolicy) {Inventory\\ Policy};
\node[block, right=4.2cm of Simulator, yshift=-1.2cm] (RoutingPolicy) {Routing\\Policy};
\node[anchor=north, yshift=0.8cm, xshift=0cm] at (InventoryPolicy.north)(Dual) {Dual Policy};
\node[block, draw, fit={(InventoryPolicy) (RoutingPolicy)(Dual)}, inner sep=0.3cm, fill opacity=0] (PolicyEvaluation) {};

\node[block, right=9cm of Simulator] (OnlineRebalance) {Online\\Rebalancing};

\node [label=center:\large Reward, yshift=0.3cm, xshift=1.8cm] at (Simulator.east) (RewardI) {};
\node [label=center:\large State, yshift=-1cm, xshift=-0.5cm] at (Simulator.south) (stateT) {};
\node [label=center:\large Action, yshift=1.8cm, xshift=0cm] at (RewardI.north) (action) {};

\draw[line] (DemandScenarios) -- (Simulator);
\draw[line] ($(Simulator.south)+(-0cm,-0cm)$) |- ($(Decision.west)+(-0cm,-0.0cm)$);
\draw[line] ($(Simulator.east)+(-0cm,0cm)$) -- ($(PolicyEvaluation.west)$);
\draw[line] ($(Decision.east)+(-0cm,-0cm)$) -| ($(PolicyEvaluation.south)+(-0cm,-0.0cm)$);
\draw[line] (PolicyEvaluation) -- (OnlineRebalance);

\draw[line] ($(PolicyEvaluation.north)+(-0cm,0cm)$)|- ($(Simulator.north)+(0cm,2cm)$) -- ($(Simulator.north)$) ;

\draw[line] (DemandScenarios) -- ($(DemandScenarios.north)+(0,4cm)$) -| ($(OnlineRebalance.north)+(0,0cm)$);



\node[anchor=north east, yshift=2.8cm, xshift=0.9cm] at (InventoryPolicy.north east) (offline){Offline Learning};

\begin{scope}[on background layer]
\node[background, fit=(Simulator)(action)(RewardI)(Decision)(InventoryPolicy)(offline)] (background) {};
\end{scope}

\end{tikzpicture}}
  }
\caption{DPRL Pipeline for Real-time Rebalancing}
\label{fig: rf}
\end{figure}
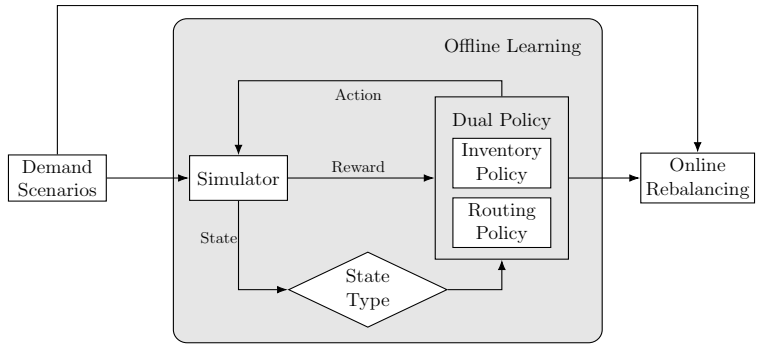

The proposed DPRL framework for the real-time rebalancing problem is illustrated in Figure~\ref{fig: rf}. During the offline learning phase, a simulator computes inventory and routing rewards and updates system states based on actions taken by the agents and demand from various demand scenarios. The rewards, system states, state type, and actions contribute to training the dual policy for both inventory and routing policies. Once the offline learning is complete, the refined dual policy is applied to the online rebalancing phase, ensuring efficient bike allocation to meet demand fluctuations. This phase involves deploying the policies on a test set of various demand scenarios to evaluate its performance.


\textbf{BSS Simulator.}
To ensure a realistic approximation of the reward function and accurate updates to system states, an event-driven simulator is used as an interactive environment, where events (rebalancing operations, rentals, and returns) are executed under the first-arrive-first-serve rule. A rental is fulfilled if the origin station has at least one available bike. The associated return for this successful rental is then scheduled for its arrival time at the destination station. The rental is marked as lost demand if the origin station is empty. 
Similarly, a return is fulfilled if a free dock is available at the destination station. Otherwise, the bike is redirected to the nearest station with an available dock, marking the return as lost demand. Notably, lost returns only occur if the corresponding rentals were successful. Rebalancing operations occur concurrently with customer trips and depend on vehicle arrival times at stations. These operations are executed sequentially, adhering to the chronological order of events.

\textbf{Dual DQN.}
To derive an optimal policy for dynamic rebalancing, two DQNs are employed to jointly train both the inventory decision network and the routing decision network. DQN, an off-policy Temporal-Difference (TD) RL technique, estimates the value function and predicts the expected return for specific actions in given states, thereby guiding agents toward more rewarding outcomes. Given the shared characteristics of inventory and routing decision-making in dynamic rebalancing, we utilize a similar neural network architecture for both Q-value networks, differing only in their output layers. The input layer aligns with the dimensions of state observations defined in Section~\ref{mmdp defin}. Subsequently, two fully connected dense layers are followed with Rectified Linear Unit (ReLU) activation functions. The output layer is tailored to the action space of each network, allowing the network to generate Q-values for all possible actions based on a given state. Specifically, the action space of the inventory network comprises three fill levels, while the action space of the routing network encompasses the number of stations (see Section~\ref{mmdp defin}).


\textbf{Heuristic Initialization.}
During training, we employ a heuristic rule as the initial routing policy to accelerate the dual DQN training process. Upon departure from a station, heuristic routing assigns a relatively full station to a relatively empty vehicle and vice versa, while also considering the distance between stations. We denote a routing distribution $u(n)$ to represent the probability of station $n$ being chosen when vehicle $i$ departs from its current station $x$ at step $k$. This distribution is defined as:
\begin{align}
\label{routing_heur}
u(n) = \alpha \rho_{1}(n) + (1-\alpha) \rho_{2}(n),
\end{align}
\begin{align}
\text{with} \quad \rho_{1}(n) &= (1/D_{x, n})^m/\sum_{n'}  (1/D_{x, n'})^m \label{distance} \\
\text{and} \quad \rho_{2}(n) &= [g(n)]^m/\sum_{n'} [g(n')]^m, \quad g(n) = \frac{C_{n}-d_{k}^{n}}{C_{n}} \cdot \frac{p_{k}^{i}}{\hat{C}_{i}} + \frac{d_{k}^{n}}{C_{n}} \cdot \frac{\hat{C}_{i}- p_{k}^{i}}{\hat{C}_{i}}.  \label{capacity}
\end{align}
In these equations, $\rho_{1}(n)$ represents the normalized and exponentiated inverse distance between station $n$ and the departing station $x$. This ensures higher probabilities for closer stations, aligning with the intuitive preference for nearby travel. Moreover, $\rho_{2}(n)$ indicates the relative inventory level of station $n$ with respect to the vehicle $i$. As outlined in Equation (\ref{capacity}), the function $g(n)$ computes the proportion of docks/bikes available at station~$n$ relative to its total capacity, aligning with the proportion of the vehicle's inventory level that could potentially be used by dropping off or picking up bikes at station $n$. If vehicle $i$ is relatively full, the relatively empty stations will have a higher probability $\rho_{2}(n)$, making them more likely to be selected, and vice versa. The weight $\alpha$ adjusts the influence of these two factors, enabling us to focus on different aspects of the rebalancing problem. Notably, when $m \rightarrow 0$, the distribution tends towards random routing with uniform probabilities, while $m= \infty$ implements a greedy policy favoring the best choice according to these metrics.

\section{Computational Experiments}
\label{sec:experiments}

\subsection{Dataset}
\label{sec:dataset}

Although real-world trip data is available, we opt for synthetic instances due to several considerations. Real-world data lacks information on unobserved demand, contains inconsistencies or noise, and omits operator rebalancing operations, making station inventories unreliable. To address these issues, we use trip data instances based on varying weather conditions and temporal features, allowing RL models to learn a rebalancing policy for diverse demand scenarios. We consider two ground-truth datasets, GT1 and GT2, each with a 60-station network and varying station distributions. In GT1, nine stations are located within a single city center, whereas GT2 features twelve stations across two city centers. City center stations have $40$ docks, while those outside have $20$ docks. The configuration of GT2 allows for a more even distribution of work-related trips (as individuals commute between stations in and outside city centers), resulting in less stressed trip demand. Using these two datasets is important for algorithm validation, as it reflects realistic urban commuting patterns. For each dataset, we generate 150 days with detailed trip information (origin station, departure time, destination station, and arrival time). The first 100 days in each dataset are used for training, while the remaining 50 days serve as a test set. Four vehicles, each with a capacity of 40 bikes, are available for rebalancing the stations.

\subsection{Benchmarks}
\label{sec:baselines}

We evaluate our DPRL algorithm algorithm against various MIP and RL models. 
\begin{itemize}
\item \textbf{Static Rebalancing (SR):} This model uses the MIP approach from \cite{liang2024dynamic} to optimize initial station inventories for the beginning of the day. The test days simulate user demand starting from these optimized initial inventories, without further rebalancing during the day. This static solution also establishes the initial inventory input for all other models.

\item \textbf{Dynamic Rebalancing (DR):} This method employs the multi-period dynamic rebalancing MIP model from \cite{liang2024dynamic}, with time periods of 30 minutes (\textbf{DR30}) and 60 minutes (\textbf{DR60}). It generates rebalancing strategies, indicating the number of bikes each vehicle should rebalance at which station for each time period, based on average rentals and returns from the training set. The rebalancing strategy obtained is also evaluated on the test set. 

\item \textbf{RL Inventory with Heuristic Routing (RIHR):}
In this approach, an MMDP model is dedicated to learning an inventory policy upon a vehicle's arrival at a station. Routing decisions upon departure are managed by a heuristic that selects the next station, as detailed in Section~\ref{rlframe} with $m=\infty$. 

\item \textbf{RL Routing with Heuristic Inventory (RRHI):} An MMDP model is trained solely on routing policy to determine the next station to visit upon vehicle departure. The target level, representing the ideal inventory level of bikes that ensures optimal service, serves as the heuristic inventory rule. Upon arriving at a station, a vehicle endeavors to rebalance the station inventory to this target level. We utilize the target level in \cite{liang2023dynamic}, calculated based on demand prediction considering weather and temporal factors. 

\item \textbf{RL with Simultaneous Inventory and Routing decisions (RSIR):} This model from \cite{liang2024reinforcement} integrates inventory and routing decisions at the moment of a vehicle's arrival at a station. 
    
\end{itemize}

All the RL models mentioned above are evaluated on the test set after training.

The MIP models are solved using IBM ILOG CPLEX v20.1.0.0 on 2.70 GHz Intel Xeon Gold 6258R machines with 8 cores, terminating when the optimization gap falls below 0.01\% or after 24 hours. The RL models are trained on a single GPU Tesla V100-PCIE-32GB. More details on the hyperparameter settings are provided in Appendix \ref{appendix_hyper}.

\subsection{Results}
\label{sec:results}

We consider a planning horizon of 4 hours from 7 a.m. to 11 a.m., encompassing the morning peak demand period.  This period also corresponds to an episode in RL models. We compare the performance of our proposed DPRL model against various benchmark models on the test set to evaluate its practical efficacy and robustness in handling new and complex demand scenarios.

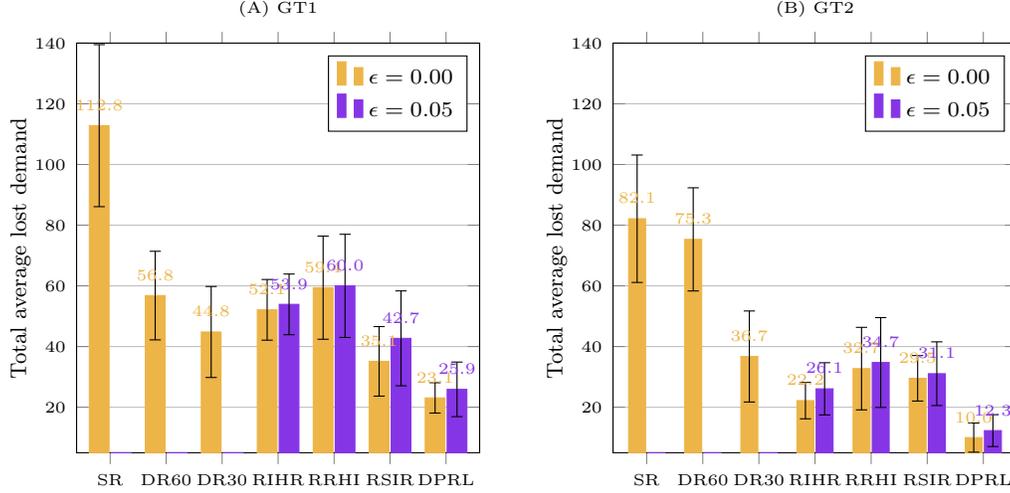
\begin{figure}[!hbtp]
    \centering
    \vspace{-\baselineskip}
      \noindent\makebox[\textwidth]{
      \resizebox{7cm}{6.7cm}{\definecolor{purple}{HTML}{8535E5}
\definecolor{yellow}{HTML}{EDB448}
\pgfplotsset{
/pgfplots/my legend/.style={
legend image code/.code={
\draw[thick,black](-0.05cm,0cm) -- (1cm,0cm);%
   }
  }
}
\begin{tikzpicture}
    \begin{axis}[
        x=0.7cm,
        width  = \textwidth,
        height = 7cm,
        title = \tiny (A) GT1,
        ybar=2*\pgflinewidth,
        bar width=7pt,
        ymajorgrids = true,
        ylabel = {\scriptsize Total average lost demand},
        symbolic x coords={A,B,C,D,E,F,G},
        xtick ={A,B,C,D,E,F,G},
        xticklabels={SR, DR60, DR30, RIHR, RRHI, RSIR, DPRL},
        scaled y ticks = false,
        enlarge x limits= 0.1, 
        ymin=5, ymax=140,
        legend pos=north east,
        nodes near coords,
        every node near coord/.append style={font=\tiny, yshift=2pt, 
        /pgf/number format/.cd,
            fixed,
            fixed zerofill,
            precision=1},
        tick label style={font=\tiny},
        ylabel near ticks, 
        ylabel shift=-6pt ]

        \addplot[
            style={yellow,fill=yellow,mark=none},
            error bars/.cd,
            y dir=both, y explicit,
            error bar style={color=black}
        ] coordinates {
            (A,112.8) +- (0,26.7)
            (B,56.8)  +- (0,14.6)
            (C,44.8)  +- (0,15)
            (D,52.1)  +- (0,10)
            (E,59.4)  +- (0,17)
            (F,35.12)  +- (0,11.46)
            (G,23.08) +- (0,5)
        };
        
        \addplot[
            style={purple,fill=purple,mark=none},
            error bars/.cd,
            y dir=both, y explicit,
            error bar style={color=black}
        ] coordinates {
            (A,0)     +- (0,0)
            (B,0)     +- (0,0)
            (C,0)     +- (0,0)
            (D,53.9)  +- (0,10)
            (E,60.0)  +- (0,17)
            (F,42.7)  +- (0,15.64)
            (G,25.89) +- (0,9)
        };

        \legend{\scriptsize $\epsilon = 0.00$, \scriptsize $\epsilon = 0.05$}
        
    \end{axis}
\end{tikzpicture}}
      \resizebox{7cm}{6.7cm}{\definecolor{purple}{HTML}{8535E5}
\definecolor{yellow}{HTML}{EDB448}

\pgfplotsset{
/pgfplots/my legend/.style={
legend image code/.code={
\draw[thick,black](-0.05cm,0cm) -- (1cm,0cm);%
   }
  }
}
\begin{tikzpicture}
\begin{axis}[
        x=0.7cm,
        width  = \textwidth,
        height = 7cm,
        title = \tiny (B) GT2,
        ybar=2*\pgflinewidth,
        bar width=6pt,
        ymajorgrids = true,
        ylabel = {\scriptsize Total average lost demand},
        symbolic x coords={A,B,C,D,E,F,G},
        xtick ={A,B,C,D,E,F,G},
        xticklabels={SR, DR60, DR30, RIHR, RRHI, RSIR, DPRL},
        scaled y ticks = false,
        enlarge x limits= 0.1, 
        ymin=5, ymax=140,
        legend pos=north east,
        nodes near coords,
        every node near coord/.append style={font=\tiny, yshift=2pt, 
        /pgf/number format/.cd,
            fixed,
            fixed zerofill,
            precision=1},
        tick label style={font=\tiny},
        ylabel near ticks, 
        ylabel shift=-6pt ]

        \addplot[
            style={yellow,fill=yellow,mark=none},
            error bars/.cd,
            y dir=both, y explicit,
            error bar style={color=black}
        ] coordinates {
            (A,82.12)  +- (0,21)
            (B,75.32)  +- (0,17)
            (C,36.72)  +- (0,15)
            (D,22.18)  +- (0,6)
            (E,32.73)  +- (0,13.6)
            (F,29.54)  +- (0,7.5)
            (G,9.99)   +- (0,4.8)
        };
        
        \addplot[
            style={purple,fill=purple,mark=none},
            error bars/.cd,
            y dir=both, y explicit,
            error bar style={color=black}
        ] coordinates {
            (A,0)     +- (0,0)
            (B,0)     +- (0,0)
            (C,0)     +- (0,0)
            (D,26.07) +- (0,8.6)
            (E,34.74) +- (0,14.8)
            (F,31.05) +- (0,10.5)
            (G,12.295)+- (0,5.25)
        };

        \legend{\scriptsize $\epsilon = 0.00$, \scriptsize $\epsilon = 0.05$}
        
    \end{axis}
\end{tikzpicture}}
      }
    \vspace{-\baselineskip}
    \caption{Total averge lost demand on test set for GT1 and GT2 }
    \label{fig:mainresults}
\end{figure}

Figure~\ref{fig:mainresults} presents the average episodic lost demand and its standard deviation, on the test set, for our algorithm and benchmarks across both GT1 and GT2. We examine two settings for RL models, distinguished by the values of $\epsilon$, which denotes the probability of selecting a random action. When $\epsilon = 0$, actions are chosen deterministically based on the highest Q-value from the trained network.  To also evaluate the robustness of the RL models, we set $\epsilon = 0.05$ to introduce a degree of randomness in action selection. This exploration can reveal more effective strategies that might have been missed during training and help prevent overfitting by occasionally deviating from the deterministic policy. Since MIP models do not have $\epsilon$, their performance is only shown for $\epsilon = 0$. 


As observed in these experiments, the SR model consistently incurs the highest lost demand, with 112.82 for GT1 and 82.1 for GT2, highlighting its inability to adapt to dynamic changes throughout the day. In contrast, dynamic rebalancing strategies, such as DR30 and DR60, improve significantly over SR by incorporating user demand information throughout the day. However, they still underperform compared to some RL models.

Among the RL models, RSIR performs competitively against RIHR and RRHI, as it integrates both inventory and routing decisions.
Nonetheless, RSIR is limited by its concurrent decision-making approach, which does not account for system changes during vehicle inventory operations. Our DPRL model demonstrates superior adaptability and effectiveness in minimizing lost demand, achieving the lowest lost demand values of 23.1 for GT1 and 10.0 for GT2. Specifically, DPRL reduces lost demand by 48.4\% for GT1 and 72.8\% for GT2 compared to DR30, and by 34.2\% for GT1 and 66.1\% for GT2 compared to RSIR. By decoupling inventory and routing decisions, DPRL highlights the effectiveness of a more realistic and granular framework for DBRP. For further details, the results of RIHR with different values of $m$ and $\alpha$ are provided in Appendix~\ref{appendix_routing_heur}. In all cases, these methods underperform compared to DPRL.



\begin{figure}[!htbp]
    \centering
    \begin{minipage}{0.48\linewidth}
        \centering
        \includegraphics[width=\linewidth]{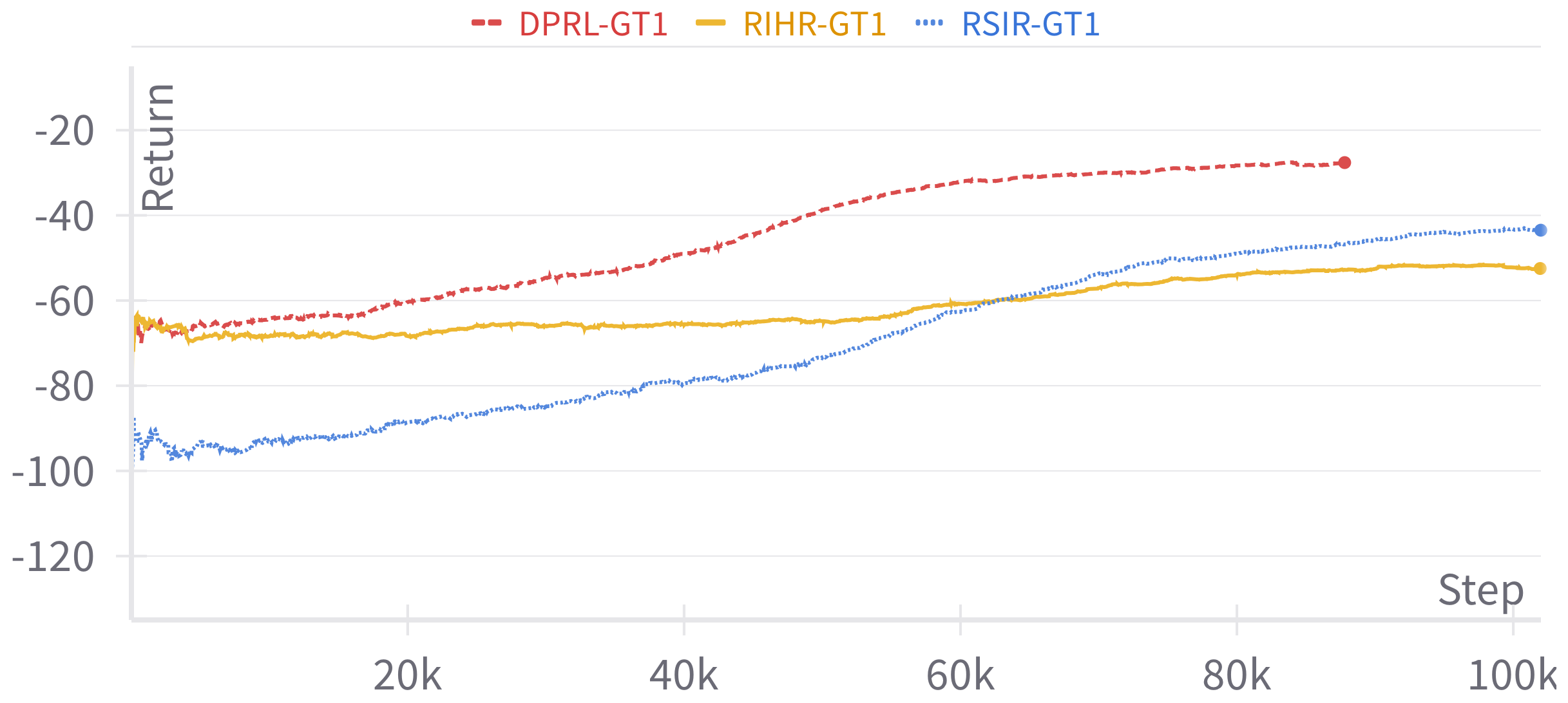}
    \end{minipage}
    \hfill
    \begin{minipage}{0.48\linewidth}
        \centering
        \includegraphics[width=\linewidth]{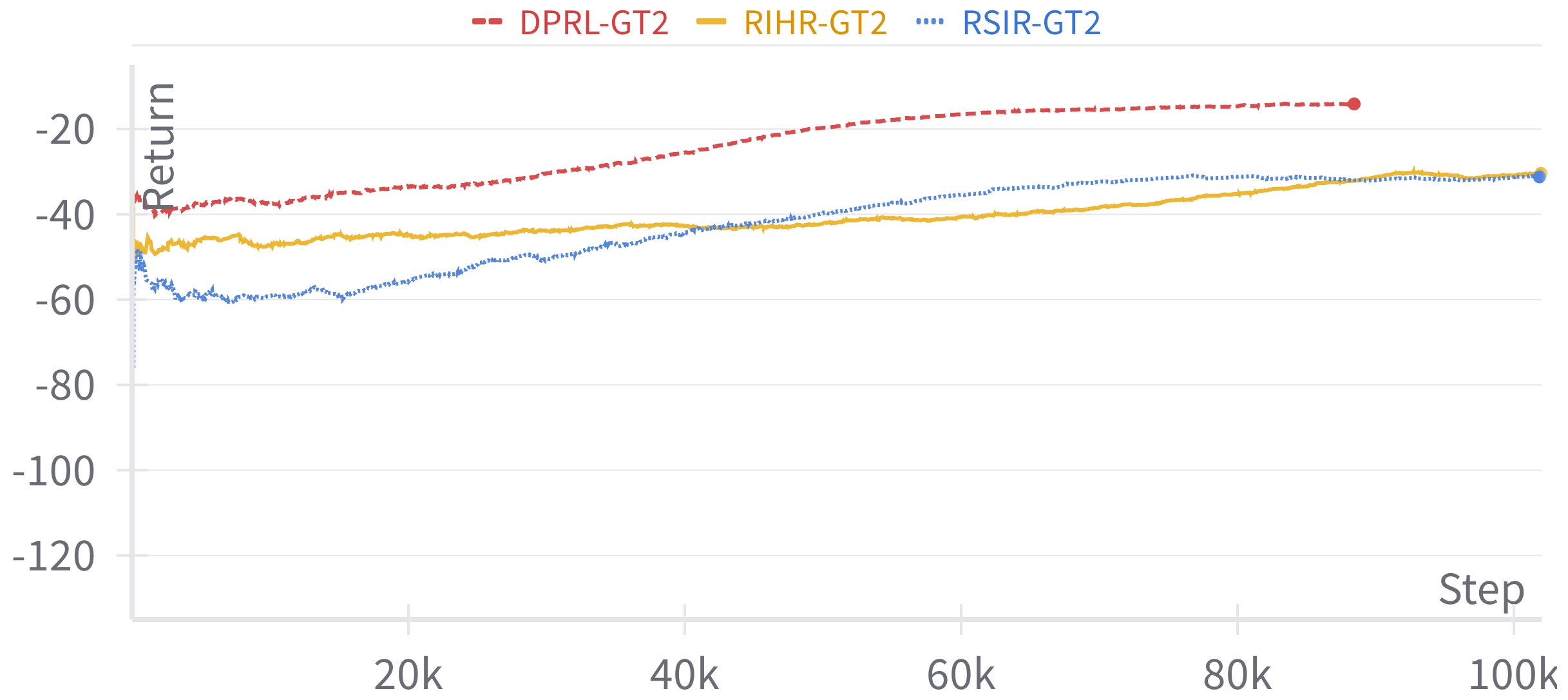}
    \end{minipage}
    \caption{Episodic return during RL training for GT1 and GT2}
    \label{fig:return}
\end{figure}

To comprehensively understand the learning process, Figure~\ref{fig:return} presents the training episodic return of three high-performing RL models. Each step represents one episode, corresponding to our planning horizon. The episodic return quantifies the cumulative reward achieved per episode, with higher values indicating more successful rebalancing strategies.

Figure~\ref{fig:return} shows a general upward trend in episodic return across all RL models, indicating improved cumulative rewards as training progresses. Notably, the DPRL model demonstrates a more consistent and steeper increase in episodic returns than the other two models. This suggests that DPRL develops a more effective rebalancing strategy over the planning horizon. Additional details regarding TD loss and Q-value are provided in Appendix~\ref{appendix_learn}. 

\subsection{Ablation Analysis}
\label{sec:ablation}

In this section, we conduct an ablation study to examine the impact of different training initializations. Additional ablation studies focusing on network architecture and heuristic routing are detailed in Appendix~\ref{appendix_archi} and Appendix~\ref{appendix_routing_heur}, respectively. The following results are based on the GT1 dataset. We vary the parameter $m$ in Equations~\eqref{distance}~and~\eqref{capacity} to initialize the training process. Table~\ref{lostdemand} presents the test set's evaluation results. Note that $m = \infty$ represents a greedy heuristic routing rule, where the fullest vehicles are assigned to the emptiest stations, and vice versa.

\begin{table}[!tbhp]
\centering
\caption{Total average lost demand (mean $\pm$ standard deviation) on test set with varying $m$}
\label{lostdemand}
\scalebox{0.85}{
\begin{tabular}{l|ll}
\hline
Initialization with $m$ & $\epsilon = 0.00$ & $\epsilon = 0.05$ \\ \hline
$m=0$                   & $19.66   \pm 6.51$       & $28.03  \pm 14.84$           \\
$m=1$                   & $23.08 \pm 5.04$            &$25.89 \pm 9.12$           \\
$m=2$                   & $33.86 \pm 7.99$                 & $40.71 \pm 13.56$              \\
$m=\infty$              & $25.53  \pm 8.54$           & $31.02  \pm 11.83$      \\ \hline    
\end{tabular}}
\end{table}

When $m=0$, the initialization for heuristic routing leads to a random selection of stations, yielding the lowest lost demand of 19.66 with $\epsilon = 0.00$. This uniform selection may prevent overfitting to specific station characteristics, leading to a highly effective rebalancing solution. However, its lack of robustness is evident as lost demand increases significantly when randomness is introduced ($\epsilon = 0.05$). For $m=1$, the heuristic routing balances distance and inventory level, resulting in lost demands of 23.08 and 25.89 for $\epsilon = 0.00$ and $\epsilon = 0.05$, respectively. This balanced approach provides flexibility and robustness, making it relatively efficient for rebalancing. For $m=2$, the increased sensitivity to both distance and inventory levels leads to a higher lost demand of 33.86, suggesting that excessive prioritization of these factors may cause overfitting and reduce effectiveness in a dynamic environment. Finally, when $m=\infty$, the strict routing rule addresses extreme imbalances but lacks the nuanced approach required for more subtle rebalancing scenarios.

\section{Conclusions}
\label{sec:conclusion}

This study introduced a novel approach for addressing real-time rebalancing problems in BSS using a dual policy RL framework. By decoupling inventory and routing decisions, our approach offered enhanced realism and efficiency over previous approaches that simultaneously made these decisions without considering system changes in detail. We modeled the inventory and routing subproblems as an MMDP within a continuous time framework. To facilitate learning, we developed a comprehensive simulator under the first-arrive-first-serve rule, enabling the computation of immediate rewards and state updates across diverse demand scenarios. 

We conducted extensive experiments and ablation studies on diverse datasets, demonstrating the superior performance of our dual policy framework compared to several benchmarks. Overall, the proposed framework significantly reduced lost demand, providing valuable insights for BSS operators and contributing to more intelligent and resilient solutions for real-world dynamic programming problems.

The research perspectives connected to our work are numerous. A first research avenue involves scaling up our models to larger BSS networks and considering the specificities of E-bike sharing systems (due to charging decisions). Additional model extensions can also include real-time traffic information and more sophisticated user behavior models. Finally, the methodological insights learned in this work are broad in their applicability and could be further applied to other balancing problems occurring in (possibly autonomous) mobility-on-demand \citep[see, e.g.][]{Jungel2023}.

\bibliographystyle{IEEEtranSN}
\bibliography{main.bib} 


\appendix
\section{Appendix}

\subsection{Hyperparameters of RL Set-up}
\label{appendix_hyper}

We introduce the hyperparameters used in our dual policy algorithm in Table~\ref{tab:para}.

\begin{table}[!tbhp]
\centering
\caption{Parameters in training process}
\label{tab:para}
\scalebox{0.85}{
\begin{tabular}{l|r}
\hline
Parameters                  & Values    \\ \hline
Total time step             & 3,000,000   \\
Learning rate               & 2.5e-4    \\
Buffer size                 & 10,000     \\
Discount factor $\gamma$                    & 0.99      \\
Batch size                  & 256       \\
Exploration rate $\epsilon$ & 1 $\rightarrow$ 0.05 \\
Exploration fraction        & 0.5       \\
1st layer neuron            & 1,024      \\
2nd layer neuron            & 512      \\ \hline
\end{tabular}}
\end{table}

\subsection{Learning Process of RL models}
\label{appendix_learn}

\begin{figure}[!htbp]
    \centering
    \begin{minipage}{0.48\linewidth}
        \centering
        \includegraphics[width=\linewidth]{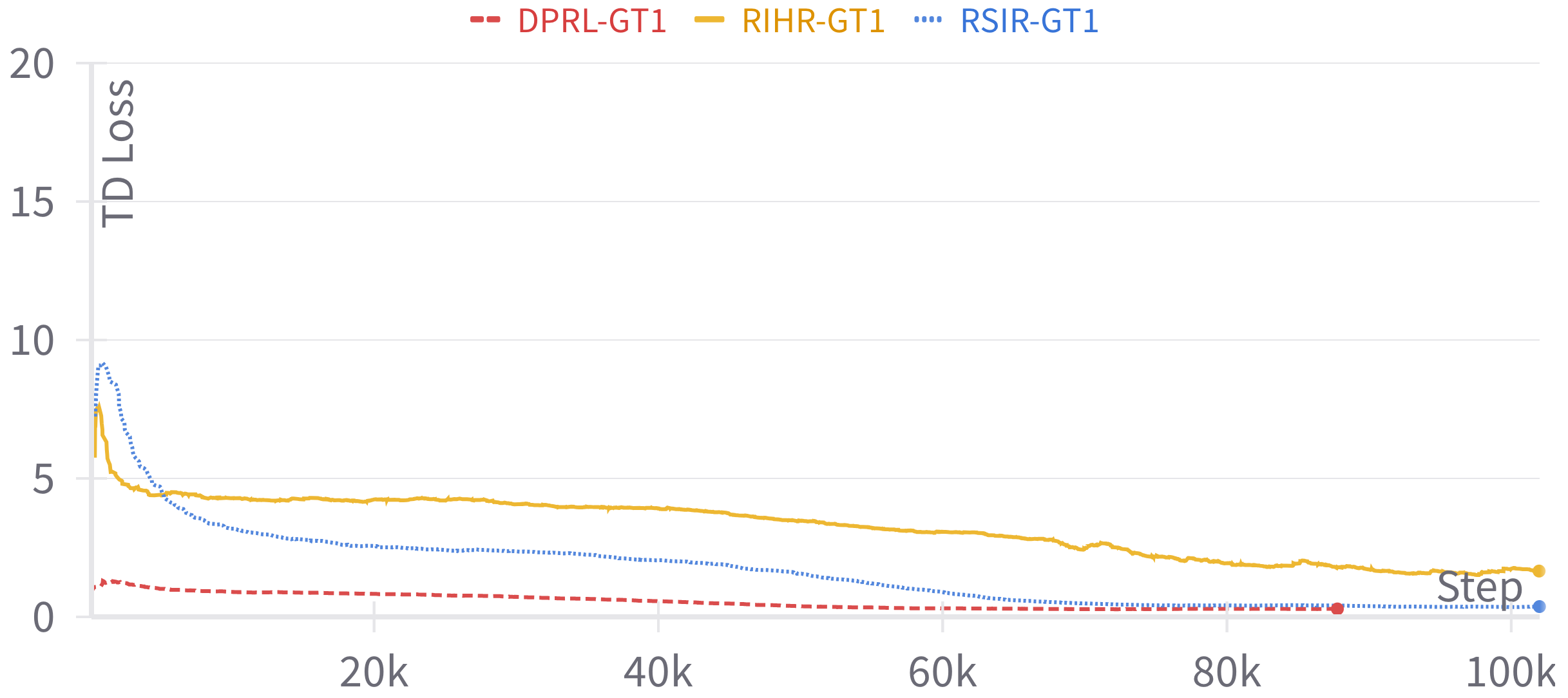}
    \end{minipage}
    \hfill
    \begin{minipage}{0.48\linewidth}
        \centering
        \includegraphics[width=\linewidth]{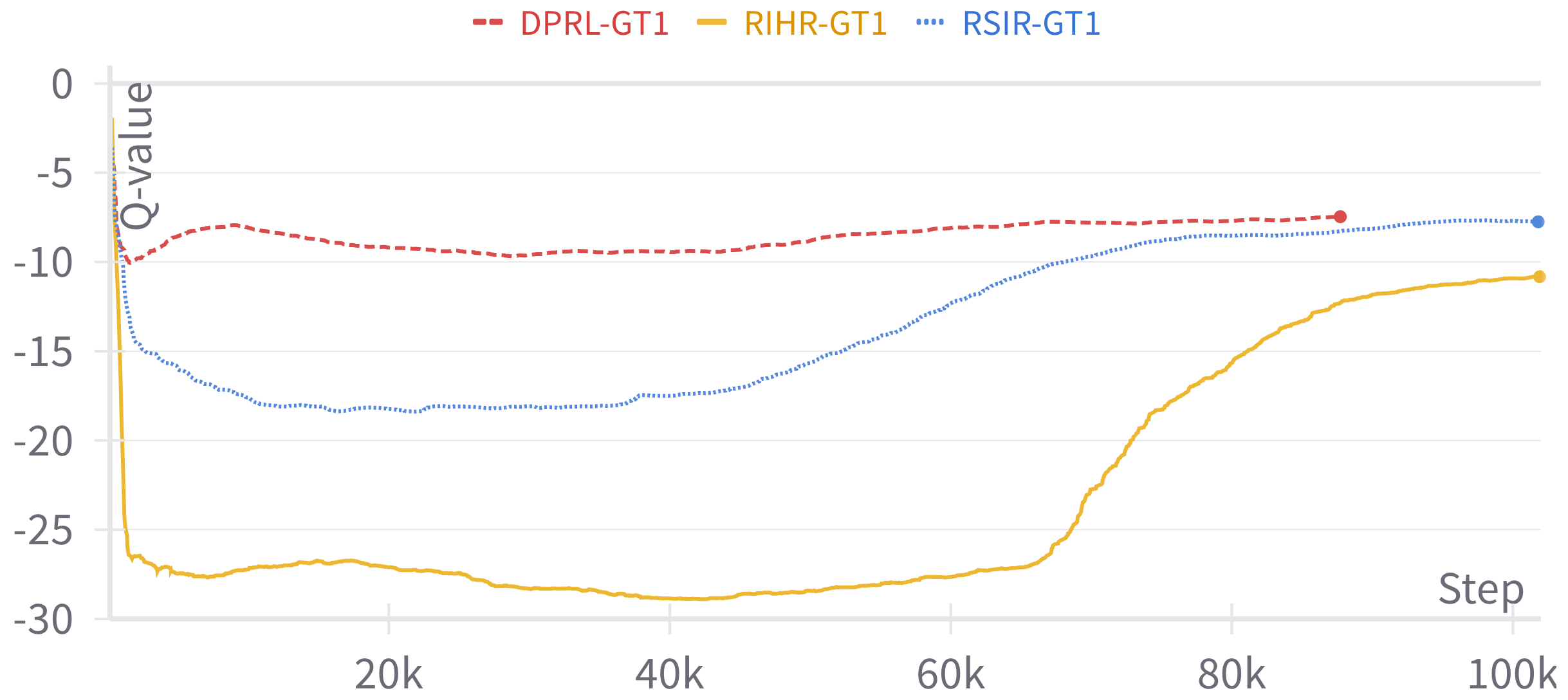}
    \end{minipage}
    \caption{TD loss and Q-value during RL training for GT1}
    \label{fig:GT1Train}
\end{figure}

\begin{figure}[!htbp]
    \centering
    \begin{minipage}{0.48\linewidth}
        \centering
        \includegraphics[width=\linewidth]{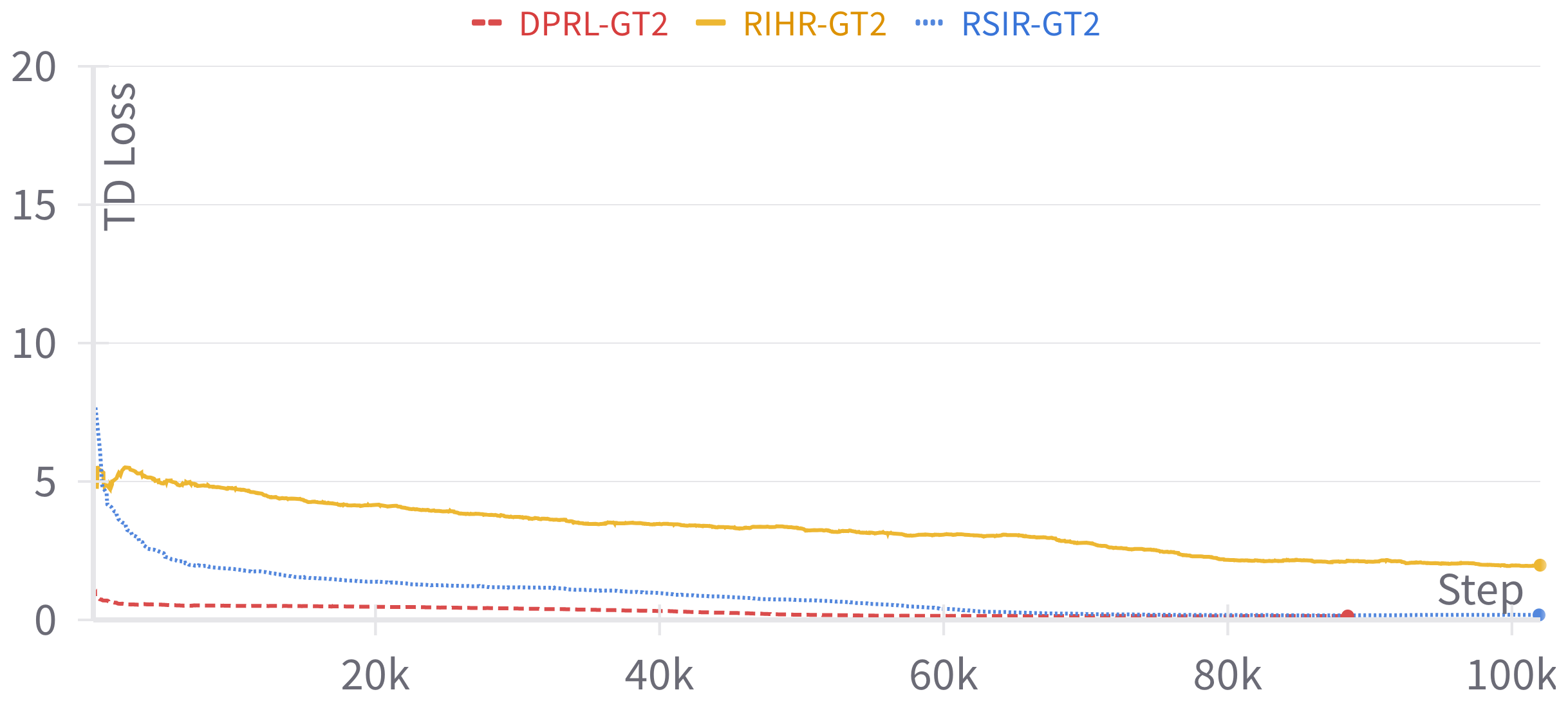}
    \end{minipage}
    \hfill
    \begin{minipage}{0.48\linewidth}
        \centering
        \includegraphics[width=\linewidth]{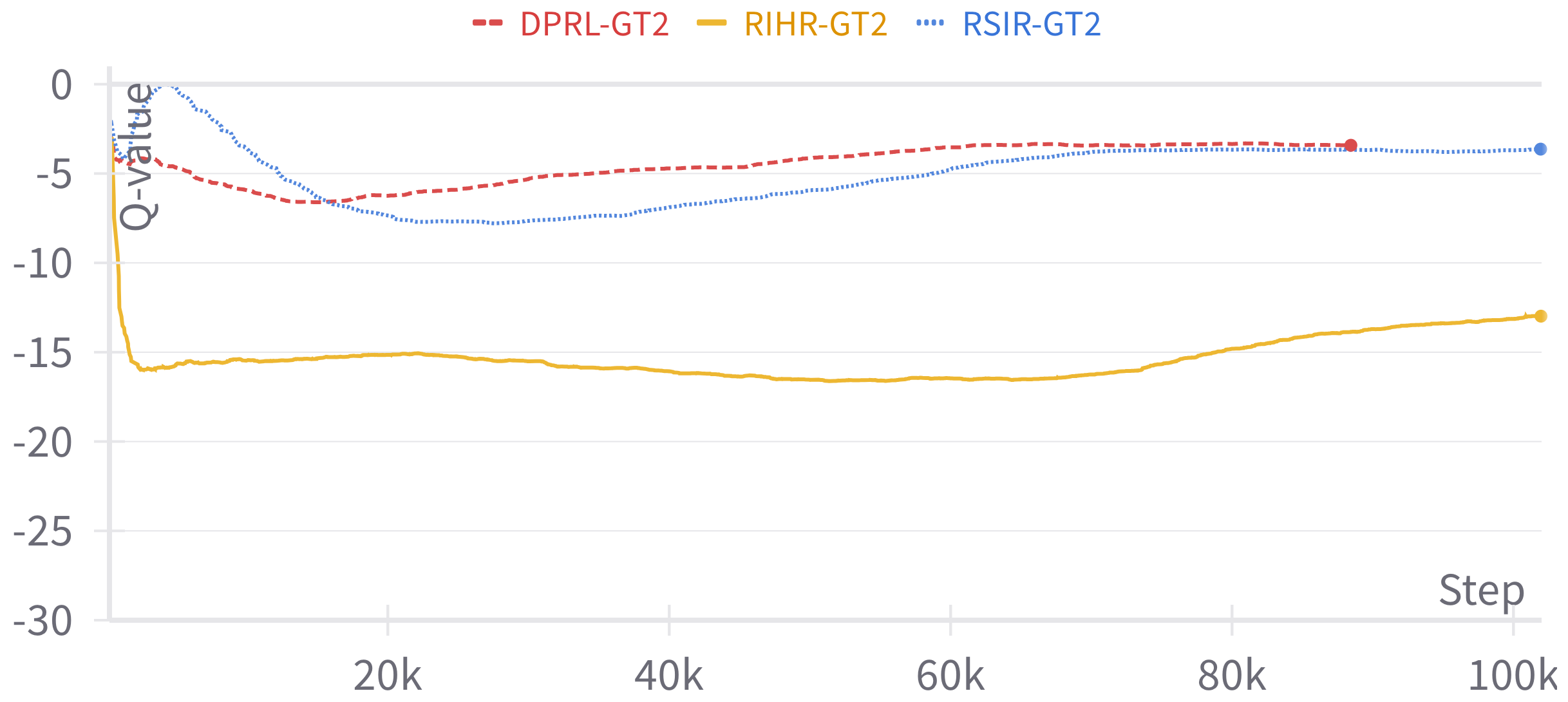}
    \end{minipage}
    \caption{TD loss and Q-value during RL training for GT2}
    \label{fig:GT2train}
\end{figure}

For a comprehensive understanding of the algorithms' performance, we report the TD Loss and Q-value during training for both GT1 and GT2 in Figure~\ref{fig:GT1Train} and Figure~\ref{fig:GT2train}. TD Loss measures the divergence between the predicted and the target Q-values as an indicator of the network’s precision in forecasting future rewards. Q-value represents the expected reward for taking a particular action in a state and following the policy thereafter. An increase in Q-value over time signifies that, as training progresses, higher quality actions are selected, reducing expected lost demand. 

The TD Loss trends downward for all RL models, indicating overall learning improvement. DPRL maintains a consistently low TD Loss, suggesting faster and more accurate Q-value estimation. Initially, the Q-value of all RL models decreases as agents explore the state-action space with a random policy. As agents learn from these experiences, Q-values increase steadily, leading to a better rebalancing policy. DPRL converges to a higher Q-value, demonstrating its ability to guide the agent towards optimizing behavior for maximum cumulative rewards.
.

\subsection{Ablation Study of Policy Architecture}
\label{appendix_archi}

\begin{figure}[!htbp]
    \centering
    \begin{minipage}{0.48\linewidth}
        \centering
        \includegraphics[width=\linewidth]{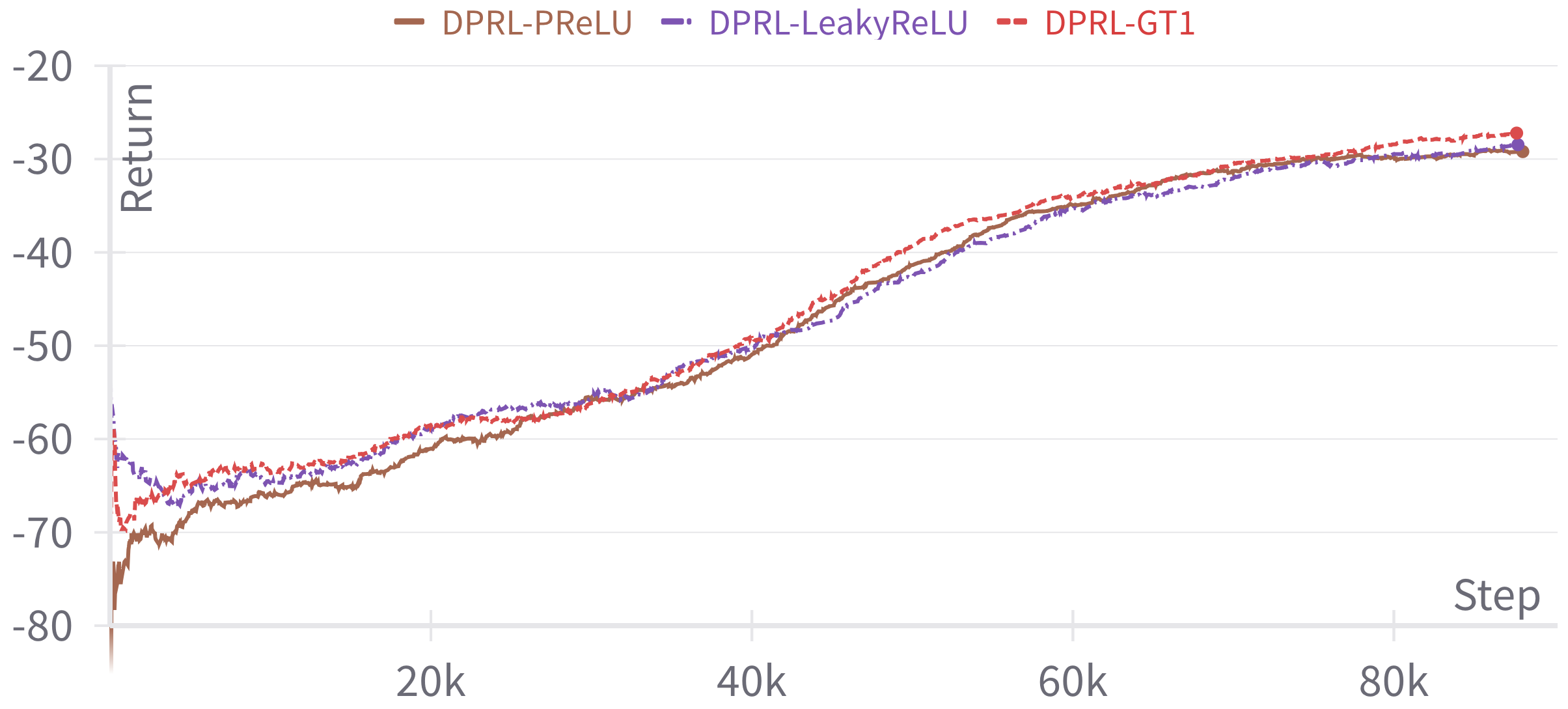}
    \end{minipage}
    \hfill
    \begin{minipage}{0.48\linewidth}
        \centering
        \includegraphics[width=\linewidth]{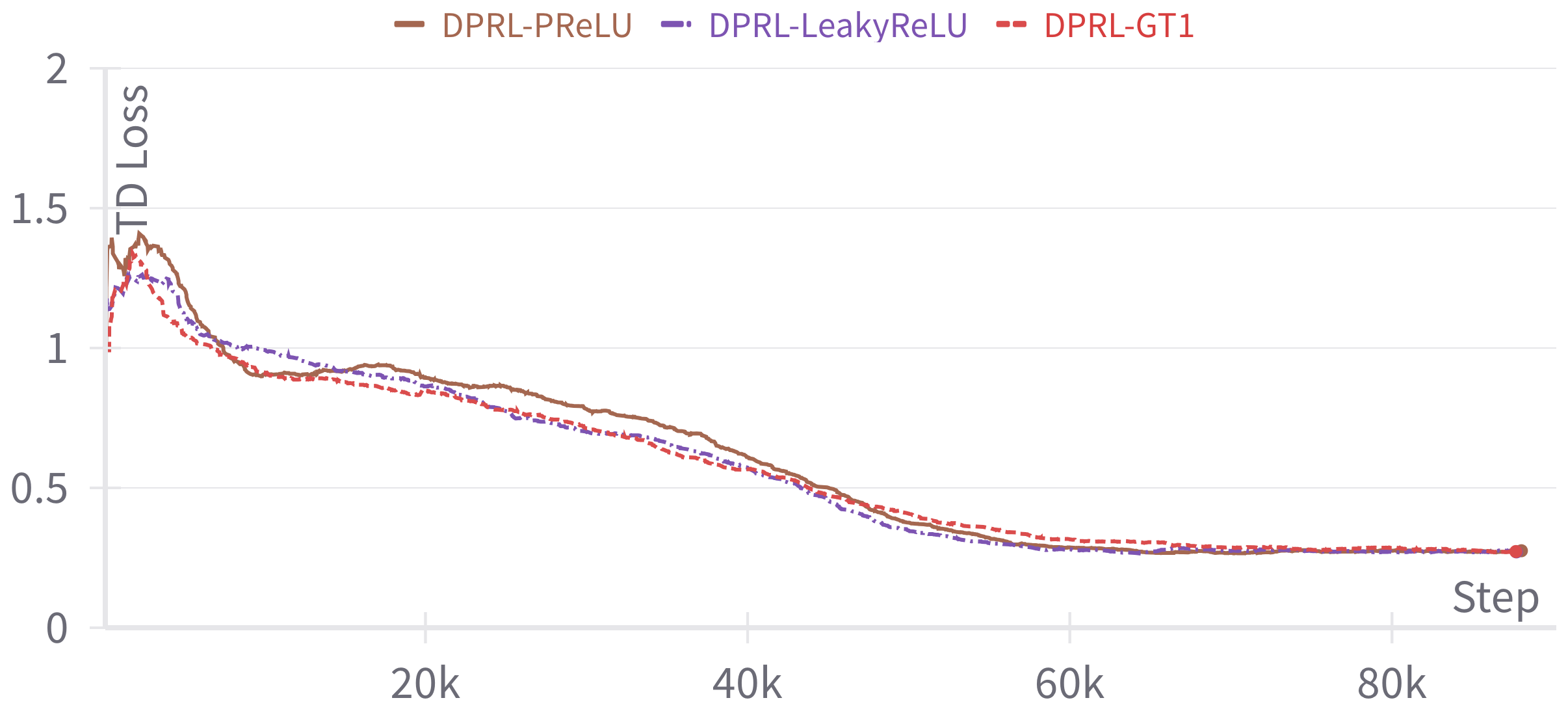}
    \end{minipage}
    \caption{Episodic return and TD loss for DPRL with different activation functions}
    \label{fig:AF}
\end{figure}

We first investigate the impacts of activation functions in the output layer. As the immediate reward in DPRL is defined as the negative value of the lost demand, we focus on Leaky ReLU and Parameterized ReLU (PReLU) due to their ability to handle negative outputs \citep{pedamonti2018comparison}. The evaluation results and training process are depicted in Table ~\ref{lostdemand_activation} and Figure~\ref{fig:AF}.

\begin{table}[!tbhp]
\centering
\caption{Total average lost demand on test set of DPRL with different activation functions}
\label{lostdemand_activation}
\scalebox{0.85}{
\begin{tabular}{l|rr}
\hline
Activation Function                                                 & $\epsilon = 0.00$ & $\epsilon = 0.05$ \\ \hline
DPRL-LeakyReLU     &     22.16        &      29.16        \\
DPRL-PReLU     &   21.73          &    27.45       \\DPRL     &   23.08          &    25.89       \\ \hline        
\end{tabular}}
\end{table}

Overall, the performance differences among the three RL models are not significant. For $\epsilon = 0.00$, both DPRL-LeakyReLU and DPRL-PReLU slightly reduce lost demand compared to the original DPRL, which has a lost demand of 23.08. However, when $\epsilon = 0.05$, DPRL shows superior performance, achieving the lowest lost demand. while DPRL-LeakyReLU and DPRL-PReLU are slightly effective in scenarios with less randomness, their performance may degrade with increased exploration. The inherent flexibility of having no activation function might help navigate the state-action space more effectively in highly exploratory scenarios.

According to Figure~\ref{fig:AF}, the episodic return and TD loss during training are very similar for these three RL models, indicating comparable learning dynamics across all activation functions.

\begin{figure}[!htbp]
    \centering
    \begin{minipage}{0.48\linewidth}
        \centering
        \includegraphics[width=\linewidth]{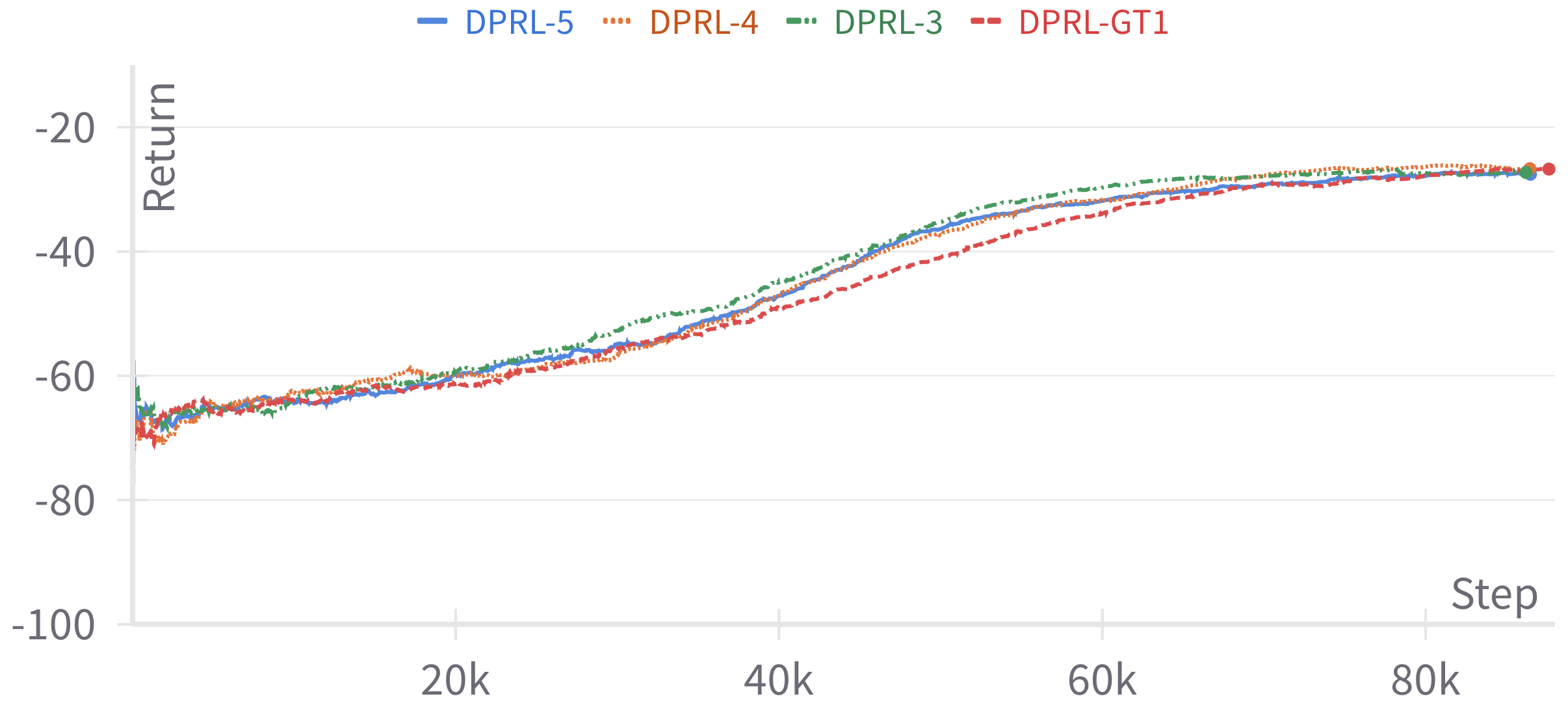}
    \end{minipage}
    \hfill
    \begin{minipage}{0.48\linewidth}
        \centering
        \includegraphics[width=\linewidth]{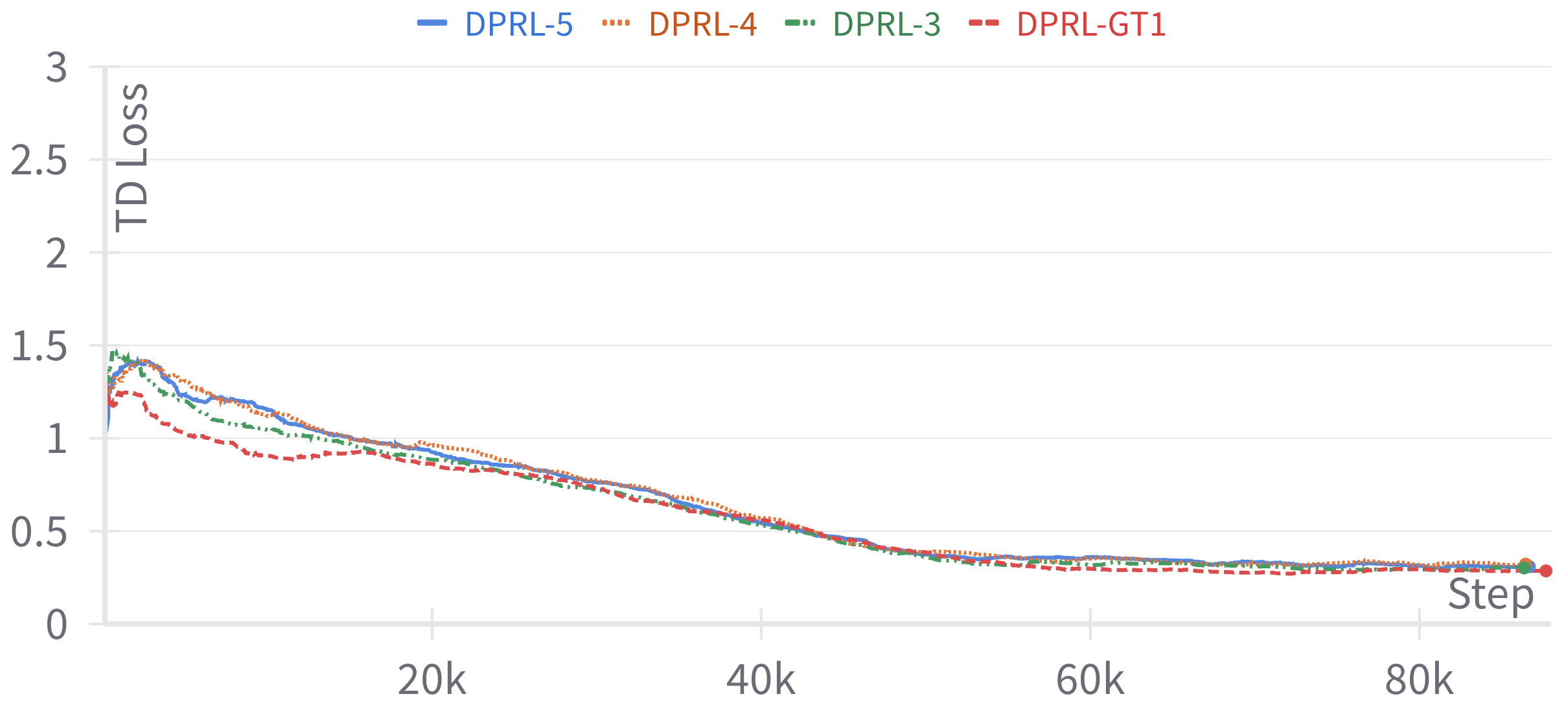}
    \end{minipage}
    \caption{Episodic return and TD loss for DPRL with different layers}
    \label{fig:AL}
\end{figure}

We then investigate the impact of different numbers of layers. As in Section~\ref{rlframe}, our DPRL consists of two fully connected dense layers following the input layer. Here, we test variants with three, four, and five layers, denoted as DPRL-3, DPRL-4, and DPRL-5, respectively. The average lost demand on test set is reported in Table~\ref{lostdemand_al}. The episodic return and TD loss during training are illustrated in Figure~\ref{fig:AL}.

\begin{table}[!tbhp]
\centering
\caption{Total average lost demand on test set of DPRL with different number of policy network layers}
\label{lostdemand_al}
\scalebox{0.85}{
\begin{tabular}{l|rr}
\hline
\# of Layers                                                 & $\epsilon = 0.00$ & $\epsilon = 0.05$ \\ \hline
DPRL-3     &     19.43       &      26.21        \\
DPRL-4    &   22.03          &    25.45       \\DPRL-5     &   22.25          &    26.70       \\ \hline        
\end{tabular}}
\end{table}

From Table~\ref{lostdemand_al}, adding layers yields only marginal improvements in performance. The additional layers allow the model to capture more complex representations, thereby capturing intricate patterns in the data. However, this improvement comes at the expense of increased model complexity. A higher number of parameters may lead to overfitting. Notably, when $\epsilon = 0.05$, the additional layers may hinder exploration and result in underperformance for most variants. Thus, there exists a trade-off between complexity and performance.

From Figure~\ref{fig:AL}, all variants exhibit similar episodic return, suggesting that deeper layers do not significantly impact the overall learning process. However, DPRL consistently achieves the lowest TD loss, indicating more accurate value estimation.

\subsection{Ablation Study of the Routing Heuristic}
\label{appendix_routing_heur}
In this section, we report an ablation study of the routing heuristic by varying its parameters $m$ and $\alpha$.

We first fix $m =1$ and obtain the evaluation on test set for varying $\alpha$, as shown in Table~\ref{lostdemand_alpha}.

\begin{table}[!tbhp]
\centering
\caption{Total average lost demand on test set with RL Inventory policy with heuristic routing by varying $\alpha$}
\label{lostdemand_alpha}
\scalebox{0.85}{
\begin{tabular}{l|l|rr}
\hline
Model                                                  & $\alpha$ & $\epsilon = 0.00$ & $\epsilon = 0.05$ \\ \hline
\multirow{5}{*}{Inventory policy \& heuristic routing} & 0.0      & 66.32             & 67.87             \\
                                                       & 0.1      & 67.26             & 67.59             \\
                                                       & 0.5      & 67.80             & 69.42             \\
                                                       & 0.8      & 66.81             & 67.23             \\
                                                       & 1.0      & 68.37             & 67.76    \\ \hline        
\end{tabular}}
\end{table}

When $\alpha = 0$, the lost demand is 66.32, indicating that relying solely on inventory level without factoring in distance results in relatively low lost demand since it aligns with our objective. As $\alpha$ increases, the lost demand slightly rises, suggesting that the introduction of distance begins to influence routing decisions but not significantly. When $\alpha = 0.8$, the lost demand slightly decreases to 66.81, showing that giving more weight to distance starts to improve rebalancing effectiveness, as vehicles are likely to visit more stations due to the relatively short distance between two consecutive stations visited in the planning horizon. Finally, when $\alpha = 1.0$, the lost demand increases again to 68.37, indicating that relying solely on distance makes vehicles visit only nearby stations without considering the actual demand among stations. However, they all underperform our DPRL significantly, underscoring the importance of learning both inventory and routing decisions.

We then explore the influence of $m$ on a single-policy RL model, where the inventory policy is trained alongside heuristic routing. The results are demonstrated in Table~\ref{lostdemand_m}. This differs from our DPRL algorithm, as illustrated in Table~\ref{lostdemand}.

\begin{table}[!tbhp]
\centering
\caption{Total average lost demand on test set with RL Inventory policy with heuristic routing by varying $m$}
\label{lostdemand_m}
\scalebox{0.85}{
\begin{tabular}{l|l|rr}
\hline
Model                                                  & $m$      & $\epsilon = 0.00$ & $\epsilon = 0.05$ \\ \hline
\multirow{4}{*}{Inventory policy \& heuristic routing} & 0        & 66.64             & 70.31             \\
                                                       & 1        & 66.81             & 67.23             \\
                                                       & 2        & 59.10             & 60.29             \\
                                                       & $\infty$ & 52.12             & 53.87   \\ \hline         
\end{tabular}}
\end{table}

When $m=0$, representing a random selection in the heuristic routing, we observe a lost demand of 66.64. This lack of prioritization can result in suboptimal rebalancing due to the absence of a structured decision-making process. Unlike our DPRL initialization with $m=0$, which can potentially benefit from random selection through learning, the fixed routing rule here won't have the opportunity to adapt and improve over time. As $m$ increases to 2, the lost demand decreases, suggesting that amplifying the effect of both distance and inventory level in heuristic routing may enhance the effectiveness of the routing policy by more aggressively prioritizing stations with significant imbalances. For $m=\infty$, corresponding to a strict assignment of the emptiest station to the fullest vehicle and vice versa, we observe a decrease in lost demand to 52.12. However, while strict prioritization based on extreme inventory levels can improve the results when only the inventory policy is learned through RL and routing decisions are given by a heuristic rule, it underperforms DPRL, where both inventory and routing are learned through RL.


\end{document}